\documentclass[sigconf]{acmart}
\usepackage{wrapfig}
\usepackage[utf8]{inputenc}

\usepackage[toc, acronym]{glossaries}
\usepackage{array}
\usepackage{xspace}
\makeglossaries
\renewcommand\footnotetextcopyrightpermission[1]{}
\AtBeginDocument{%
  }

\newacronym{psf}{PSF}{Point-Spread Function}
\newacronym{mlp}{MLP}{Multi Layer Perceptron}
\newacronym{inr}{INR}{Implicit Neural Representation}

\newcommand{\pspace}{projection space\xspace}
\newcommand{\tspace}{target space\xspace}
\newcommand{\RTR}{\textit{\textbf{Ray2Ray}}\xspace}

\begin{document}

\title{Efficient Proxy Raytracer for Optical Systems using 
Implicit Neural Representations}

\author{Shiva Sinaei\textsuperscript{*}}
\email{shiasinaei98@gmail.com}
\orcid{0009-0002-0837-5125}

\affiliation{%
  \institution{University of Osaka}
  \city{Osaka}
  \state{}
  \country{Japan}
}
\author{Chuanjun Zheng}
\email{chuanjunzhengcs@gmail.com}
\orcid{0009-0005-7211-380X}
\affiliation{%
  \institution{University College London}
  \city{London}
  \country{United Kingdom}
}

\author{Kaan Ak\c{s}it}
\email{k.aksit@ucl.ac.uk}
\orcid{0000-0002-5934-5500}
\affiliation{%
  \institution{University College London}
  \city{London}
  \country{United Kingdom}
}

\author{Daisuke Iwai}
\email{daisuke.iwai.es@osaka-u.ac.jp}
\orcid{0000-0002-3493-5635}
\affiliation{%
  \institution{University of Osaka}
  \city{Osaka}
  \country{Japan}
}

\begin{abstract}
Ray tracing is a widely used technique for modeling optical systems, involving 
sequential surface-by-surface computations which can be computationally 
intensive.
We propose \RTR, a novel method that leverages implicit neural representations to 
model optical systems with greater efficiency, eliminating the need for 
surface-by-surface computations in a single pass end-to-end model.
\RTR learns the mapping between rays emitted from a given source and their 
corresponding rays after passing through a given optical system in a physically 
accurate manner.
We train \RTR on nine off-the-shelf optical systems, acheiving 
positional errors on the order of \(1\,\mu\mathrm{m}\) and angular
deviations on the order \(0.01\) degrees in the estimated output rays.
Our work highlights the potentials of neural representations as a proxy optical 
raytracer.
\end{abstract}

  

\maketitle


\begin{figure}
  \hspace{1cm} 
  \includegraphics[width=0.95\linewidth]{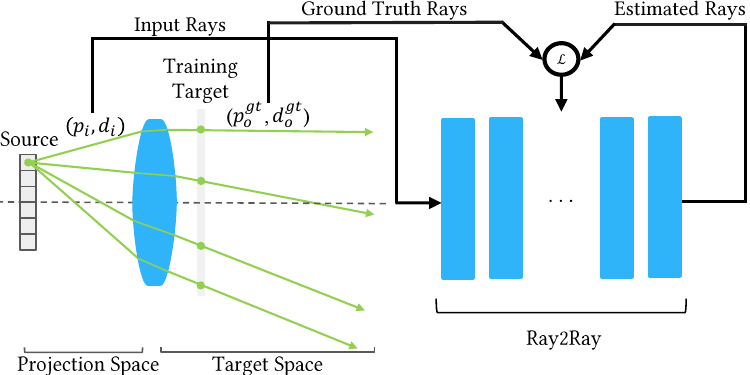} 
  \caption{Ray tracing through an optical system (left), Our proposed INR-based proxy ray tracer, \RTR(right).}
  \label{fig:concept}
\end{figure}
\section{Introduction}

Ray tracing is a well-established method for simulating light propagation in 
optical systems, enabling precise modeling of the transformations induced by optical 
systems on incident light.
Ray tracing process typically involves a series of geometric intersections
and refractions, computed using Newton's method and Snell's law, 
respectively~\citep{wang2022differentiable}. 
However, the iterative nature of Newton's method can lead to increased computational 
load in ray tracing, particularly when tracing a large number of rays or dealing with 
more complex optical configurations.

Recently, neural networks have emerged as a powerful alternative for modeling the 
ray tracing through optical systems, offering a promising substitute for traditional 
ray tracing. 
%
%
For instance, \citep{tseng2021differentiable} propose a hybrid architecture that 
combines a \gls{mlp} with convolutional layers, mapping incoming light from infinity, 
parameterized by its incident angle, to a \gls{psf}. Primarily aimed at optical system 
optimization, their model is conditioned on the parameters of the optical system.
Similarly, \citep{yang2023aberration} employ a pure \gls{mlp} that maps discrete 
3D points within a specific frustum in \tspace to a \gls{psf}, conditioned on the 
system's focus point. 
Existing works mainly focus on modeling the ray tracing through optical systems within an 
imaging-based regime, where the transformations introduced by the optical system are 
represented as a \gls{psf} generated from a point in \tspace. 
However, this approach limits the supported active area in \tspace and 
limits the generalizability of the model to the specific application it was designed for.
In contrast, we approach this task from a 
projection-based perspective, adopting a ray-based model that lifts the imposed 
limitations. 

The proposed \RTR is a learned model that learns the effect of the optical system as
transformation of rays emitted from a given source to their corresponding 
rays on a target plane, as they propagate through the optical system.
\RTR is a ray-based model that preserves both the spatial origin and directional 
information of 
rays, enabling the simulation of optical system effects at arbitrary depths in \tspace 
and eliminating the need to repeatedly query the 
model for targets at different distances.
Additionally, the ray-based nature of \RTR enhances generalizability, as it does not 
freeze the light to 
a predefined representation. Unlike \gls{psf}-based models which are limited by the 
number of rays and the kernel size used during their training, \RTR supports queries with 
an arbitrary number of rays and can be applied at any resolution.
Furthermore, by adopting a projection-based formulation, \RTR allows for more accurate 
estimation of the optical system's effects for a given source, without the need for 
dense sampling of points in \tspace, which limits imaging-based approaches to 
partition-wise \gls{psf}s that may not accurately represent the effects across the entire 
target plane. Our key contribution is summarized as follows:
\begin{itemize}
  \item Learned Proxy Ray Tracer. We propose a lightweight model serving as a 
  proxy for the ray 
  tracing of optical systems. It learns ray propagation using an 
  Implicit Neural Representation (INR)
  based on ray parameters. Our model achieves high accuracy in predicting 
  the output rays from the optical system, attaining positional errors 
  on the order of \(1\,\mu\mathrm{m}\) and angular deviations on the order of \(0.01\) degrees.

\end{itemize}

\section{Method}

We aim to develop a proxy ray tracer as an alternative to conventional ray tracing 
in optical systems 
by leveraging INRs. In the context of ray tracing, a ray 
\( R = (\mathbf{p}, \mathbf{d}) \) is typically defined by its origin \( \mathbf{p} \) 
and its 
normalized direction \( \mathbf{d} \).
Our model, \RTR, learns to map each input ray emitted from a given source 
in \pspace (\( \mathbf{p}_i, \mathbf{d}_i \)), to its corresponding output ray in \tspace 
(\( \mathbf{p}_o, \mathbf{d}_o \)), after it has propagated through the optical system.
Here, \( \mathbf{p}_i \in \mathbb{R}^2 \) and \( \mathbf{d}_i \in \mathbb{R}^2 \) represent
the ray's origin 
and direction on the source, respectively, while \( \mathbf{p}_o \in \mathbb{R}^2 \) and 
\( \mathbf{d}_o \in \mathbb{R}^2 \) 
denote the position and direction of the ray upon reaching the target plane, respectively. 
We train \RTR as following optimization problem: 
\begin{equation}
  \hat{w}_{\text{r2r}} \leftarrow \arg\min_{w_{\text{r2r}}} \mathcal{L}(\text{\RTR}(p_i, d_i), (p_o^{\text{gt}}, d_o^{\text{gt}}))
  \label{eq:representation}
\end{equation}
where \(w_{\text{r2r}}\) encodes the parameters that implicitely capture the ray tracing 
process for the underlying optical system. In equation \eqref{eq:representation}, 
\( \mathcal{L} \) represents any valid loss function that measures the discrepancy between 
the predicted rays by \RTR and the ground truth rays \( (p_o^{\text{gt}}, d_o^{\text{gt}}) \).
Once \RTR is trained for a given optical system, it can be queried 
with rays defined in \pspace and the optical system's 
effect at any target plane in \tspace is evaluated through a simple 
ray-target intersection. An overview of our method is provided in fig. \ref{fig:concept}.

\paragraph*{\textit Synthesized Dataset.} To train \RTR for each optical system, we 
need a dataset consisting of pairs of input rays
and corresponding ground truth output rays. For simulation, we consider a 
grid of size \( M \times M \) as source. 
In this setup, \(\mathbf{p}_i\) is defined as the center position of each cell on 
our source grid
in the \pspace. 
For each cell in our grid, we generate 1024 rays, whose directions \(\mathbf{d}_i\) are calculated 
using Monte Carlo sampling at the entrance of the optical setup. All rays are simulated with a 
fixed wavelength of \(589~\mathrm{nm}\).
%
%
The rays are traced through the optical system, and their intersections with a 
target plane are computed, resulting in the output ray positions \(\mathbf{p}_o\) and 
directions \(\mathbf{d}_o\). 
In our setup, the training target plane is positioned at a close distance of $1\mathrm{mm}$ from the 
optical system in \tspace, 
which was found to improve the model's convergence during training. 
The optical system simulation was carried out using the optics toolkit \text{odak} 
\citep{Aksit_Odak}.
We allocated 80\% of the grid cells for training and the remaining 20\% for testing 
the performance of our learned model.

\paragraph*{\textit Model and Training.}
To learn the complex mapping from input rays in \pspace{} to their corresponding output
rays in \tspace{}, we employ a simple \gls{mlp} architecture that includes residual skip
connections introduced every three layers to enhance training stability. Each hidden
layer is followed by a ReLU activation function to incorporate non-linearity. To
investigate the influence of architectural design and select an appropriate 
model, we conduct an ablation study on nine different off-the-shelf optical systems, evaluating various \gls{mlp} configurations with different
numbers of layers and neurons. The results and discussion of this study are provided in the
supplementary material. The model is trained for 3000 epochs using the AdamW optimizer
with an initial learning rate of \(5 \times 10^{-4}\), which decays exponentially
throughout the training process.

\section{Results and Discussion}
We evaluate a customized optical system consisting of a single lens with an 
aperture size of 
\(50.6\,\mathrm{mm}\) and a focal length of \(60\,\mathrm{mm}\), using a source grid 
of size \(12 \times 12\,\mathrm{mm}\), divided 
into \(24 \times 24\) cells. To assess the generalization capability of 
our model, we conduct two 
experiments: one in which the learned \RTR is tested on a previously 
unseen cell of the grid source, 
and another using a novel ray pattern comprising \(1{,}000{,}000\) rays 
different from training dataset. 
The positional and angular errors for each experimental condition 
are summarized in Table~\ref{tab:generalization}.
\begin{table}[h]
  \centering
  \caption{\RTR Performance under different experimental conditions}
  \label{tab:generalization}
  \begin{tabular}{@{}lcc@{}}
  \toprule
  \textbf{Condition} & \textbf{Pos. Error (\(\mu\)m)} & \textbf{Ang. Error (degrees)} \\
  \midrule
  Unseen Grid Cell   & 2.4                   & 0.06                   \\
  Novel Ray Pattern  & 5.9                     & 0.076                    \\
  \bottomrule
  \end{tabular}
  \end{table}
Overall, we believe \RTR could benefit the designers and component manufacturers 
in the computational imaging domain by providing proxy models that improve 
simulation efficiency.

\begin{acks}
This work was supported by JST ASPIRE, Grant Number JPMJAP2404.
\end{acks}

\bibliographystyle{ACM-Reference-Format}  
\bibliography{references}  

\end{document}